\documentclass[letterpaper]{article}
\usepackage[preprint]{aaai2027}
\usepackage[hyphens]{url}
\usepackage{graphicx}
\usepackage{natbib}
\usepackage{booktabs}
\usepackage{amsmath}
\usepackage{amssymb}
\usepackage{xspace}

\newcommand{\method}{SpecFlow\xspace}
\newcommand{\minade}{\operatorname{minADE}_{20}}
\newcommand{\minfde}{\operatorname{minFDE}_{20}}
\newcommand{\secondbest}[1]{\underline{#1}}

\title{SpecFlow: Flow Matching in the Acceleration Spectrum for Multi-Agent Trajectory Forecasting}
\author{
    Ziyang Xu\textsuperscript{\rm 1,\rm 2},
    Peng Han\textsuperscript{\rm 1}\corresponding,
    Hai Zhou\textsuperscript{\rm 1}
}
\affiliations{
    \textsuperscript{\rm 1}National Space Science Center, Chinese Academy of Sciences, Beijing 100190, China\\
    \textsuperscript{\rm 2}University of Chinese Academy of Sciences, Beijing 101408, China\\
    xuziyang23@mails.ucas.ac.cn, hanpeng@nssc.ac.cn, zhouhai@nssc.ac.cn
}

\begin{document}

\maketitle

\begin{abstract}
Multimodal trajectory forecasting must cover diverse future behaviors without sacrificing dynamical coherence. Most generative forecasters evolve frame-wise positions directly, leaving the relationships among position, velocity, and acceleration implicit. We present SpecFlow, a conditional flow-matching framework that treats the Discrete Cosine Transform (DCT) coefficients of future accelerations as its generative state. SpecFlow learns a transport from Gaussian noise to acceleration spectra and reconstructs velocity and position through fixed analytical integration initialized by the last observed state. A joint phase-space objective supervises all three kinematic levels. To represent destination uncertainty without future information, an endpoint predictor produces candidate destinations and confidence scores solely from observed context, and the highest-confidence candidate conditions the spectral denoiser. Alternating attention over trajectory hypotheses and agents captures multimodality and interaction. Under best-of-20 evaluation, SpecFlow achieves average minADE/minFDE of 0.15/0.22 on ETH/UCY, 0.69/0.86 at four seconds on NBA, and 7.38/11.57 on SDD. It establishes state-of-the-art results across all five ETH/UCY scenes and achieves the lowest four-second ADE on NBA with a tied-best FDE, while remaining competitive on SDD. Together, these results demonstrate that acceleration-spectrum flow matching provides an effective structured generative state across smooth pedestrian motion, coordinated multi-agent sports dynamics, and heterogeneous outdoor trajectories. Code will be made publicly available.
\end{abstract}

\begin{figure}[t]
	\centering
	\includegraphics[width=0.65\columnwidth]{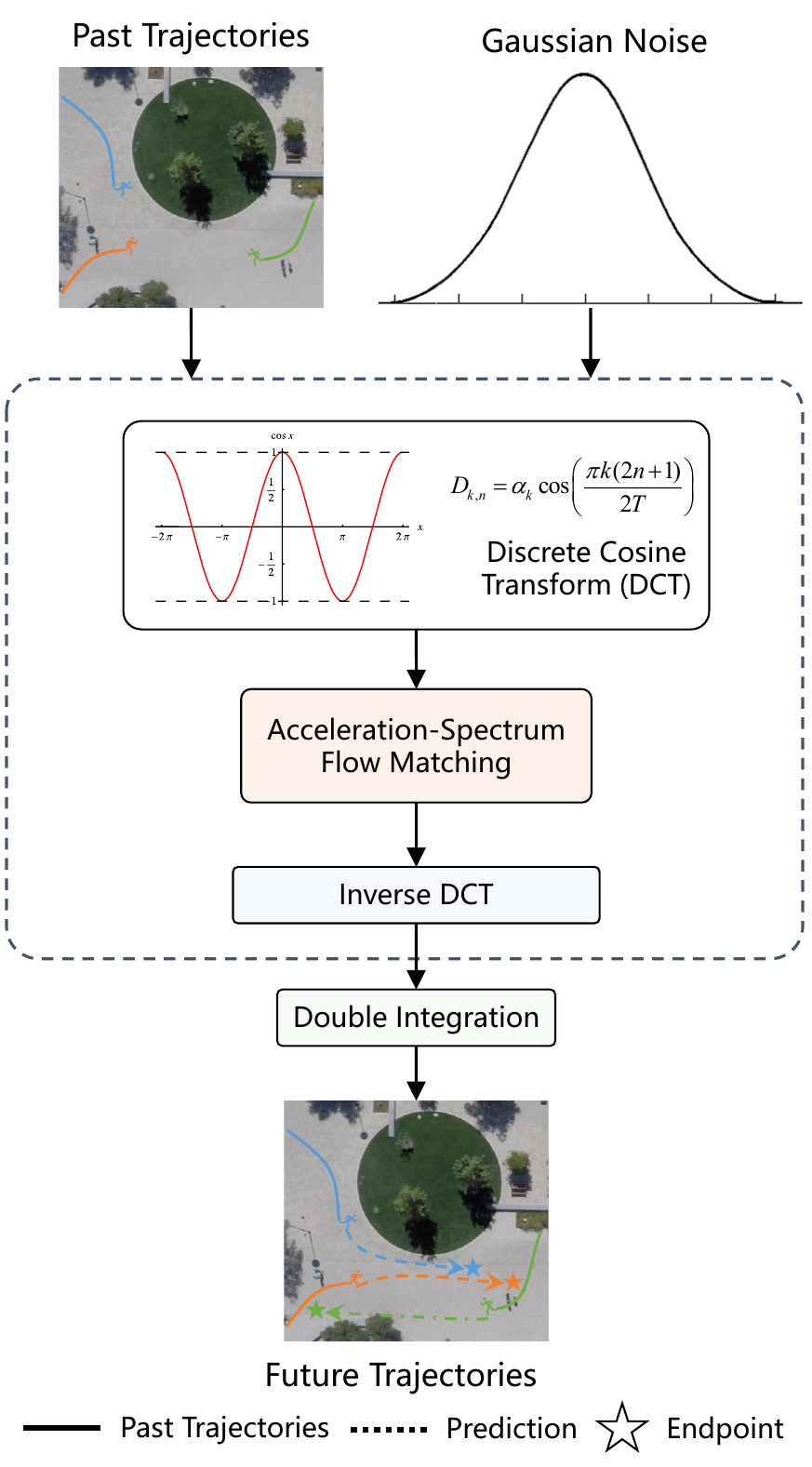}
	\caption{SpecFlow generates acceleration spectra and reconstructs future trajectories.}
	\label{fig:illustration}
	\vspace{-0.5em}
\end{figure}

\section{Introduction}

Trajectory forecasting estimates plausible future motion from observed multi-agent histories. Futures depend on scene geometry and other agents. Recurrent, graph-based, adversarial, variational, and diffusion models address this uncertainty through interaction modeling and diverse sampling~\cite{alahi2016sociallstm,gupta2018socialgan,salzmann2020trajectronpp,yuan2021agentformer,gu2022mid,mao2023led}. Flow-matching models learn a continuous transport from simple noise to the trajectory distribution~\cite{lipman2023flowmatching,fu2025moflow}. Most methods generate frame-wise positions and learn their kinematic relations implicitly.

\newpage

We define the generative state in the acceleration spectrum. The DCT provides a fixed orthonormal temporal basis ordered from low to high frequency~\cite{ahmed1974dct,mao2019dctmotion}. A network transports Gaussian noise to acceleration coefficients. Two fixed cumulative operators then recover velocity and position from the last observed state. Each acceleration sample determines all three kinematic quantities. Retaining part of the basis sets the temporal bandwidth directly.

The acceleration representation does not by itself resolve destination uncertainty or agent interaction. \method predicts endpoints from observed context and uses the selected proposal to condition the denoiser. Its backbone alternates attention along the hypothesis and agent axes. The first operation exchanges information among modes. The second models interaction within each joint forecast.

Under best-of-20 evaluation, \method averages 0.15/0.22 on ETH/UCY and ranks first in every scene. It reaches 0.69/0.86 at four seconds on NBA and 7.38/11.57 on SDD without scene inputs.

The contributions are as follows:
\begin{itemize}
  \item We formulate conditional flow matching in the DCT acceleration spectrum. Fixed analytical operators reconstruct velocity and position from the observed state.
  \item We train with a joint position--velocity--acceleration objective. Endpoint conditioning and alternating attention model destination uncertainty and agent interaction.
  \item \method leads all ETH/UCY scenes and four-second NBA ADE, ties the best NBA FDE, and remains competitive on SDD. Ablations examine the main components and sampling cost.
\end{itemize}

\section{Related Work}

\paragraph{Trajectory Prediction.}
Classical crowd models encode interaction through hand-crafted dynamics such as social forces~\cite{helbing1998social}. Social LSTM and Social GAN introduced recurrent pooling and adversarial sampling for pedestrian interaction~\cite{alahi2016sociallstm,gupta2018socialgan}. Graph and transformer models represent time-varying neighborhoods, group structure, and joint agent dependencies~\cite{mohamed2020socialstgcnn,huang2019stgat,ivanovic2019trajectron,salzmann2020trajectronpp,yuan2021agentformer,xu2022groupnet,shi2023tutr,shi2022mtr,xu2023eqmotion,kim2024highgraph}. Other work makes long-term intent explicit through anchors, endpoints, waypoints, or control points~\cite{kothari2021socialanchors,mangalam2020pecnet,mangalam2021ynet,chiara2022goaldriven,bae2023controlpoints,lin2024ppt}. IAD combines endpoint proposals with short-term polar intentions and refinement~\cite{liu2026iad}. \method also predicts endpoints, but uses them as observed-context conditions for a generative acceleration spectrum.

\paragraph{Diffusion and Flow Models.}
Latent-variable and adversarial predictors learn distributions over plausible futures~\cite{lee2017desire,gupta2018socialgan,sadeghian2019sophie,xu2022socialvae}. PCCSNet separates clustering, classification, and synthesis~\cite{sun2021pccsnet}. Diffusion models learn iterative denoising transports from Gaussian noise~\cite{ho2020ddpm}. Trajectory variants use purposive hypotheses, motion-indeterminacy diffusion, leapfrog initialization, or compact descriptors~\cite{bae2022npsn,gu2022mid,mao2023led,bae2024singular}. Flow matching regresses a continuous probability path and samples by integrating an ordinary differential equation~\cite{lipman2023flowmatching,liu2023rectifiedflow}. MoFlow applies flow matching to multi-agent trajectories and distills the transport into a one-step predictor~\cite{fu2025moflow}. \method retains an ODE sampler but transports acceleration-spectrum coefficients rather than frame-wise coordinates.

\paragraph{Spectral and Kinematic Structure.}
DCT bases provide compact representations of long temporal dependencies in human motion~\cite{ahmed1974dct,mao2019dctmotion,mao2020motionattention}. EigenTrajectory learns low-rank trajectory descriptors for multimodal forecasting~\cite{bae2023eigen}. Other models address dynamic feasibility through heterogeneous dynamics, equivariance, or explicit control points~\cite{salzmann2020trajectronpp,xu2023eqmotion,bae2023controlpoints}. \method applies the DCT to acceleration and learns the flow in the resulting coefficient space. Fixed integration recovers velocity and position.

\section{Method}
\label{sec:method}

\begin{figure*}[t]
  \centering
  \includegraphics[width=\textwidth]{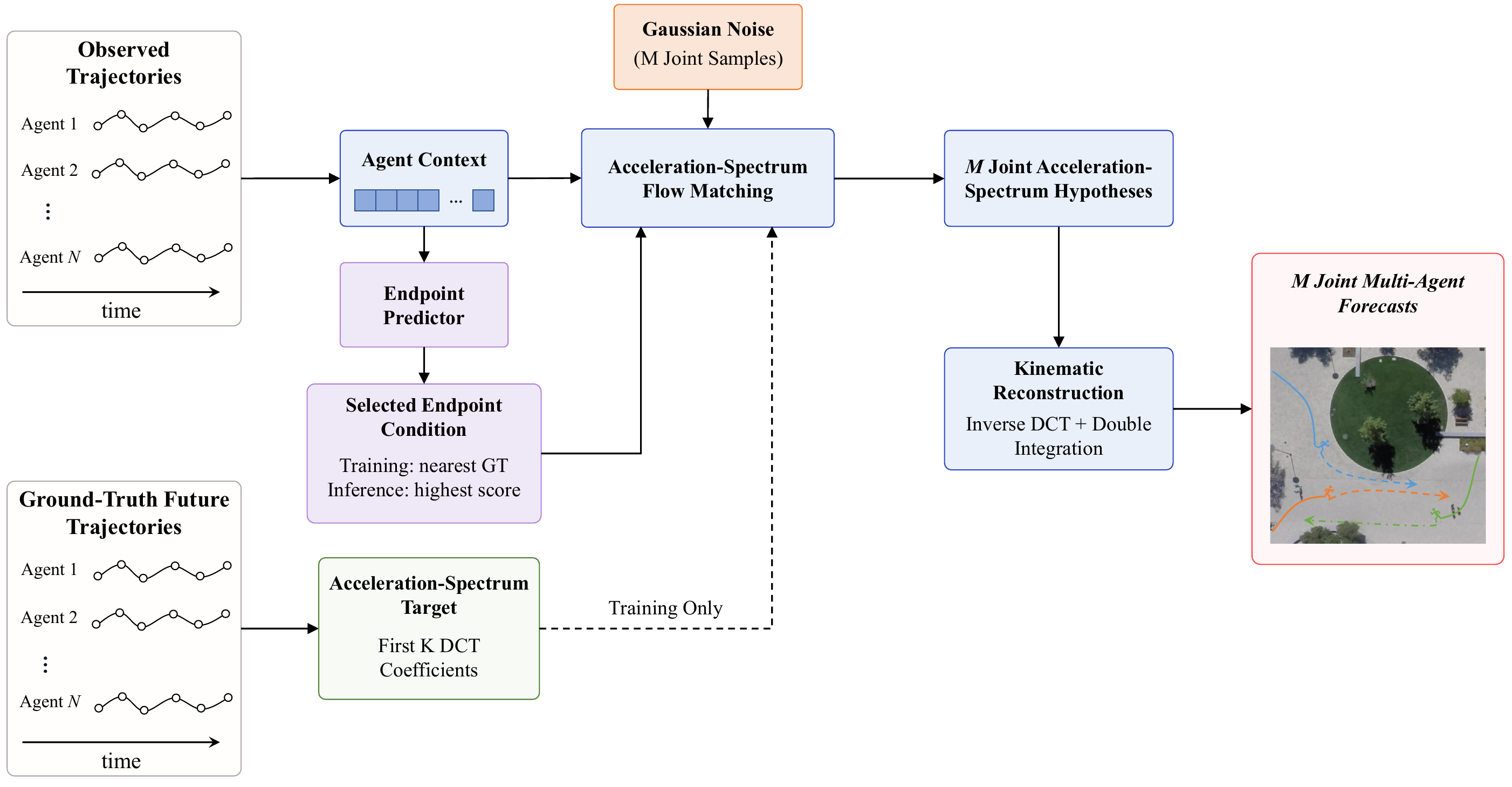}
  \caption{Observed trajectories and Gaussian noise enter the endpoint predictor and acceleration-spectrum flow model. Fixed reconstruction maps $M$ joint acceleration hypotheses to future trajectories. Ground-truth futures select the endpoint condition and spectral target only during training.}
  \label{fig:overview}
\end{figure*}

\subsection{Problem Formulation}

Let $\mathbf{X}^{o}\in\mathbb{R}^{A\times P\times 2}$ denote $P$ observed positions for $A$ agents, and let $\mathbf{X}^{f}\in\mathbb{R}^{A\times T\times 2}$ denote their future positions. Coordinates are relative to the final observation. A context encoder maps absolute position, relative position, and finite-difference velocity to agent tokens $\mathbf{H}\in\mathbb{R}^{A\times d}$. The model generates $M$ joint hypotheses $\{\widehat{\mathbf{X}}^{f,m}\}_{m=1}^{M}$. Each hypothesis forecasts all $A$ agents. We use $M=20$ in every experiment.

\subsection{Acceleration Spectrum}

Finite differences of the future sequence give velocity $\mathbf{V}^{f}$ and acceleration $\mathbf{A}^{f}$. At the temporal boundary, we replicate the last valid difference as in the implementation. For horizon $T$, the orthonormal DCT-II basis $\mathbf{D}\in\mathbb{R}^{T\times T}$ is
\begin{equation}
\begin{aligned}
D_{k,n}&=\alpha_k\cos\!\left(\frac{\pi k(2n+1)}{2T}\right),\\
\alpha_0&=T^{-1/2},\qquad \alpha_{k>0}=(2/T)^{1/2}.
\end{aligned}
\end{equation}
We retain the first $K$ rows, $\mathbf{D}_K\in\mathbb{R}^{K\times T}$, and define the generative target
\begin{equation}
\mathbf{C}_{A}=\mathbf{D}_K\mathbf{A}^{f}\in\mathbb{R}^{A\times K\times2}.
\end{equation}
The basis is fixed. The value of $K$ controls temporal bandwidth, and the learned transport operates in a $2K$-dimensional state per agent.

\subsection{Conditional Spectral Flow Matching}

For each hypothesis, we sample $\mathbf{Z}\sim\mathcal{N}(\mathbf{0},\mathbf{I})$ and $t\in(0,1)$. A linear conditional path connects noise and data,
\begin{equation}
\mathbf{C}_t=(1-t)\mathbf{Z}+t\mathbf{C}_{A}.
\end{equation}
The denoiser predicts the clean spectral endpoint $\widehat{\mathbf{C}}_{A,\theta}$ from $(\mathbf{C}_t,t,\mathbf{H},\mathbf{e})$, where $\mathbf{e}$ is an endpoint embedding. This prediction induces the velocity
\begin{equation}
\mathbf{u}_{\theta}(\mathbf{C}_t,t)=
\frac{\widehat{\mathbf{C}}_{A,\theta}-\mathbf{C}_t}{1-t}.
\end{equation}
For the linear conditional path, the target vector field is
\begin{equation}
\mathbf{u}^{\star}_t=\frac{\mathrm{d}\mathbf{C}_t}{\mathrm{d}t}
=\mathbf{C}_{A}-\mathbf{Z}
=\frac{\mathbf{C}_{A}-\mathbf{C}_t}{1-t},
\end{equation}
and the standard conditional flow-matching (CFM) objective is
\begin{equation}
\mathcal{L}_{\mathrm{CFM}}
=\mathbb{E}_{\mathbf{C}_{A},\mathbf{Z},t}
\left[\left\|\mathbf{u}_{\theta}(\mathbf{C}_t,t)
-\mathbf{u}^{\star}_t\right\|_F^2\right].
\label{eq:cfm}
\end{equation}
Under the clean-endpoint parameterization,
\begin{equation}
\mathcal{L}_{\mathrm{CFM}}
=\mathbb{E}\!\left[
\frac{\left\|\widehat{\mathbf{C}}_{A,\theta}-\mathbf{C}_{A}\right\|_F^2}
{(1-t)^2}\right],
\label{eq:cfm-data}
\end{equation}
so spectral data prediction and CFM field regression parameterize the same conditional transport.
At inference, Euler integration starts from Gaussian noise and advances the velocity field from $t=0$ to $t=1$. During training, $t$ follows a logit-normal distribution that places fewer samples near the path endpoints.

Each noisy spectrum is combined with a sinusoidal time code, observed context, a hypothesis embedding, and an agent embedding. A decoder block first applies self-attention across the $M$ hypotheses for each agent. It then attends across the $A$ agents within each hypothesis. A classification head scores the generated hypotheses.

\subsection{Observed-Context Endpoint Conditioning}

For each agent, $L=20$ learned queries cross-attend to its context token. Two multilayer perceptrons produce candidate endpoints $\{\mathbf{g}_{\ell}\}_{\ell=1}^{L}$ and confidence logits $\{s_{\ell}\}_{\ell=1}^{L}$. During training, the ground-truth destination is assigned to the nearest candidate. Endpoint regression and confidence classification supervise the predictor, and the nearest proposal conditions the denoiser. At inference, the model selects $\mathbf{g}_{\operatorname*{arg\,max}_{\ell}s_{\ell}}$. Future positions and oracle goals are never used at test time. The endpoint objective is the same on all three datasets.

\begin{figure*}[t]
\centering
\includegraphics[width=\textwidth]{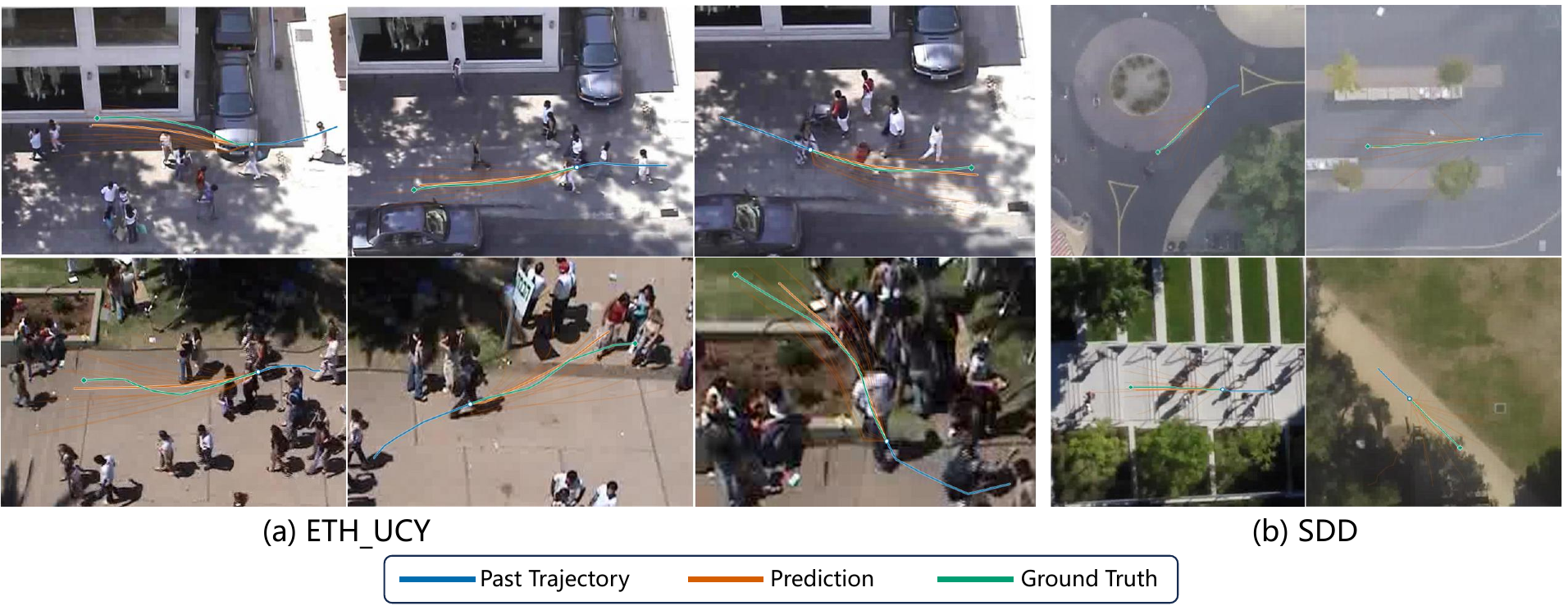}
\caption{Observed ETH/UCY and SDD trajectories are input to \method. The model outputs multiple forecasts, shown with the ground-truth futures.}
\label{fig:qualitative-eth-sdd}
\end{figure*}

\subsection{Fixed Kinematic Reconstruction}

Let $\mathbf{S}\in\mathbb{R}^{T\times T}$ be the strict lower-triangular cumulative-sum matrix, and define
\begin{equation}
\mathbf{W}_{\mathrm{int}}=\mathbf{D}_K\mathbf{S}\mathbf{D}_K^{\top},
\qquad
\mathbf{d}_{1}=\mathbf{D}_K\mathbf{1}.
\end{equation}
Given the last observed relative position $\mathbf{x}_0$ and velocity $\mathbf{v}_0$, spectral velocity and position are reconstructed as
\begin{align}
\widehat{\mathbf{C}}_{V}
&=\mathbf{W}_{\mathrm{int}}\widehat{\mathbf{C}}_{A}
  +\mathbf{d}_{1}\mathbf{v}_0^{\top},\\
\widehat{\mathbf{C}}_{X}
&=\mathbf{W}_{\mathrm{int}}\widehat{\mathbf{C}}_{V}
  +\mathbf{d}_{1}\mathbf{x}_0^{\top}.
\end{align}
The time-domain outputs are $\widehat{\mathbf{A}}=\mathbf{D}_K^{\top}\widehat{\mathbf{C}}_{A}$, $\widehat{\mathbf{V}}=\mathbf{D}_K^{\top}\widehat{\mathbf{C}}_{V}$, and $\widehat{\mathbf{X}}=\mathbf{D}_K^{\top}\widehat{\mathbf{C}}_{X}$. All matrices are precomputed and frozen. Position and velocity follow deterministically from the acceleration sample and observed initial state.

\subsection{Joint Phase-Space Objective}

For each training example, we select the nearest hypothesis in position space and supervise it at all kinematic levels:
\begin{equation}
\begin{aligned}
\mathcal{L}_{\mathrm{phase}}={}&
\lambda_x\mathcal{L}_{x}+\lambda_v\mathcal{L}_{v}
+\lambda_a\mathcal{L}_{a},\\
(\lambda_x,\lambda_v,\lambda_a)={}&(1,0.5,0.1).
\end{aligned}
\end{equation}
Each term is the mean Euclidean error over valid agents and frames. Let $\mathbf{R}_a$, $\mathbf{R}_v$, and $\mathbf{R}_x$ denote the fixed linear maps from an acceleration spectrum to acceleration, velocity, and position. Observed-state offsets cancel in prediction errors. For every phase-space level $q$,
\begin{equation}
\mathbf{R}_{q}(\widehat{\mathbf{C}}_{A}-\mathbf{C}_{A})
=(1-t)\mathbf{R}_{q}(\mathbf{u}_{\theta}-\mathbf{u}^{\star}_t).
\label{eq:phase-cfm}
\end{equation}
Thus $\mathcal{L}_{\mathrm{phase}}$ is a clean-data CFM objective under a fixed kinematic task metric and logit-normal time weighting. Its numerical value differs from the isotropic squared norm in Eq.~\eqref{eq:cfm}. The zero-error vector field is unchanged because $\lambda_a>0$ and $\mathbf{R}_a=\mathbf{D}_K^\top$ is injective on the retained spectrum. The implementation uses this structured surrogate without a separate isotropic spectral loss. The complete objective is
\begin{equation}
\mathcal{L}_{\mathrm{total}}=\mathcal{L}_{\mathrm{phase}}+
\lambda_c\mathcal{L}_{\mathrm{cls}}+
\lambda_e\mathcal{L}_{\mathrm{end}}.
\label{eq:total-loss}
\end{equation}
All losses act on the same acceleration--velocity--position reconstruction.

\section{Experiments}
\label{sec:experiments}

\subsection{Experimental Setup}

We evaluate on ETH/UCY~\cite{pellegrini2009eth,lerner2007ucy}, NBA SportVU~\cite{yue2014sportvu}, and Stanford Drone Dataset (SDD)~\cite{robicquet2016sdd}. ETH/UCY uses leave-one-scene-out evaluation with eight observed and twelve predicted frames. NBA uses ten observed frames and predicts twenty frames, or four seconds, for ten players and the ball. SDD contains 2,829 test windows with eight observed and twelve predicted frames. We report best-of-20 $\minade/\minfde$ in meters for ETH/UCY and NBA and in pixels for SDD. Baseline values retain their published precision.

\subsection{Main Results}

\paragraph{ETH/UCY.}
Table~\ref{tab:ethucy} reports the lowest minADE/minFDE pair for \method in all five scenes. Its average is 0.15/0.22, compared with second-best values of 0.19 and 0.30. UNIV has the largest FDE margin, at 0.22 versus 0.33 for HighGraph. On ETH, \method gives 0.45 FDE versus 0.51 for PPT. The FDE margins are larger than the ADE margins, placing most of the difference near the end of the prediction horizon.

The improvement is not concentrated in one unusually easy split. Using the rounded entries in Table~\ref{tab:ethucy}, \method lowers the second-best FDE by 0.06 on ETH, 0.03 on HOTEL, 0.11 on UNIV, 0.06 on ZARA1, and 0.04 on ZARA2. The relative reductions are approximately 11.8\%, 20.0\%, 33.3\%, 25.0\%, and 20.0\%. UNIV combines the largest absolute and relative gains, while HOTEL has the smallest absolute scale and still improves. This scene-wise consistency is stronger evidence for the representation than the average alone, because leave-one-scene-out evaluation changes both training composition and test geometry in every row.

\begin{table*}[t]
\centering
\small
\begin{tabular*}{\textwidth}{@{\extracolsep{\fill}}lcccccc@{}}
\toprule
Method & ETH & HOTEL & UNIV & ZARA1 & ZARA2 & AVG \\
\midrule
Social-GAN & 0.87/1.62 & 0.67/1.37 & 0.76/1.52 & 0.35/0.68 & 0.42/0.84 & 0.61/1.21 \\
Social-STGCNN & 0.64/1.11 & 0.49/0.85 & 0.44/0.79 & 0.34/0.53 & 0.30/0.48 & 0.44/0.75 \\
Social-VAE & 0.41/0.58 & 0.13/0.19 & 0.21/0.36 & 0.17/0.29 & 0.13/0.22 & 0.21/0.33 \\
MID & 0.39/0.66 & 0.13/0.22 & 0.22/0.45 & 0.17/0.30 & 0.13/0.27 & 0.21/0.38 \\
LED & 0.39/0.58 & \secondbest{0.11}/0.17 & 0.26/0.43 & 0.18/0.26 & 0.13/0.22 & 0.21/0.33 \\
EigenTrajectory & 0.36/0.53 & 0.12/0.19 & 0.24/0.43 & 0.19/0.33 & 0.14/0.24 & 0.21/0.34 \\
TUTR & 0.40/0.61 & \secondbest{0.11}/0.18 & 0.23/0.42 & 0.18/0.34 & 0.13/0.25 & 0.21/0.36 \\
HighGraph & 0.40/0.55 & 0.13/0.17 & \secondbest{0.20}/\secondbest{0.33} & 0.17/0.27 & \secondbest{0.11}/0.21 & 0.20/\secondbest{0.30} \\
PPT & 0.36/\secondbest{0.51} & \secondbest{0.11}/\secondbest{0.15} & 0.22/0.40 & 0.17/0.30 & 0.12/0.21 & 0.20/0.31 \\
MoFlow & 0.40/0.57 & \secondbest{0.11}/0.17 & 0.23/0.39 & \secondbest{0.15}/0.26 & 0.12/0.22 & 0.20/0.32 \\
IAD & \secondbest{0.34}/0.52 & 0.15/0.24 & \secondbest{0.20}/0.36 & \secondbest{0.15}/\secondbest{0.24} & \secondbest{0.11}/\secondbest{0.20} & \secondbest{0.19}/0.31 \\
\midrule
\method & \textbf{0.29/0.45} & \textbf{0.10/0.12} & \textbf{0.14/0.22} & \textbf{0.12/0.18} & \textbf{0.10/0.16} & \textbf{0.15/0.22} \\
\bottomrule
\end{tabular*}
\caption{Best-of-20 ETH/UCY results in meters. AVG is the scene mean. Baselines follow MoFlow and IAD~\cite{fu2025moflow,liu2026iad}. Bold and underlined values rank first and second.}
\label{tab:ethucy}
\end{table*}

\begin{table*}[t]
\centering
\small
\begin{tabular*}{\textwidth}{@{\extracolsep{\fill}}lcccccccc@{}}
\toprule
& \multicolumn{2}{c}{1 s} & \multicolumn{2}{c}{2 s} & \multicolumn{2}{c}{3 s} & \multicolumn{2}{c}{Total (4 s)} \\
\cmidrule(lr){2-3}\cmidrule(lr){4-5}\cmidrule(lr){6-7}\cmidrule(lr){8-9}
Method & ADE & FDE & ADE & FDE & ADE & FDE & ADE & FDE \\
\midrule
Social-GAN & 0.41 & 0.62 & 0.81 & 1.32 & 1.19 & 1.94 & 1.59 & 2.41 \\
Social-STGCNN & 0.34 & 0.48 & 0.71 & 0.94 & 1.09 & 1.77 & 1.53 & 2.26 \\
PECNet & 0.40 & 0.71 & 0.83 & 1.61 & 1.27 & 2.44 & 1.69 & 2.95 \\
NPSN & 0.35 & 0.58 & 0.68 & 1.23 & 1.01 & 1.76 & 1.31 & 1.79 \\
GroupNet & 0.26 & 0.34 & 0.49 & 0.70 & 0.73 & 1.02 & 0.96 & 1.30 \\
MID & 0.28 & 0.37 & 0.51 & 0.72 & 0.71 & 0.98 & 0.96 & 1.27 \\
LED & \textbf{0.18} & \secondbest{0.27} & 0.37 & 0.56 & 0.58 & 0.84 & 0.81 & 1.10 \\
MoFlow & \textbf{0.18} & \textbf{0.25} & 0.34 & 0.47 & 0.52 & \secondbest{0.67} & \secondbest{0.71} & \secondbest{0.87} \\
MoFlow-IMLE & \textbf{0.18} & \textbf{0.25} & 0.34 & 0.47 & 0.52 & \secondbest{0.67} & \secondbest{0.71} & \textbf{0.86} \\
IAD & \secondbest{0.21} & \textbf{0.25} & \textbf{0.32} & \textbf{0.45} & \textbf{0.48} & 0.75 & 0.79 & 0.99 \\
\midrule
\method & \textbf{0.18} & \textbf{0.25} & \secondbest{0.33} & \secondbest{0.46} & \secondbest{0.50} & \textbf{0.66} & \textbf{0.69} & \textbf{0.86} \\
\bottomrule
\end{tabular*}
\caption{Best-of-20 NBA results in meters at cumulative horizons. Baselines follow MoFlow and IAD~\cite{fu2025moflow,liu2026iad}. Bold and underlined values rank first and second.}
\label{tab:nba}
\end{table*}

\paragraph{NBA.}
Table~\ref{tab:nba} places \method among the best two entries for all eight NBA metrics. It ties the best ADE and FDE at one second. At two seconds, IAD is lower by 0.01 on both measures. At three seconds, \method has a 0.01 lower FDE, while IAD has a 0.02 lower ADE. At four seconds, \method reaches 0.69/0.86 while forecasting all 11 entities jointly. This is the lowest ADE and a tied-best FDE. The ranking changes across horizons rather than showing uniform dominance.

The horizon-wise pattern distinguishes short-term fit from long-term stability. At one second, \method, MoFlow, and MoFlow-IMLE are indistinguishable after rounding. By four seconds, \method lowers ADE from 0.71 to 0.69 relative to both MoFlow variants and matches the best FDE of 0.86. IAD is strongest at two-second ADE/FDE, whereas its FDE rises from 0.45 to 0.99 over the remaining two seconds. The corresponding increase for \method is from 0.46 to 0.86. These results suggest that the spectral acceleration state is most useful as errors accumulate over a longer rollout, rather than as a uniform gain at every intermediate horizon.

\begin{figure}[t]
\centering
\includegraphics[width=0.82\columnwidth]{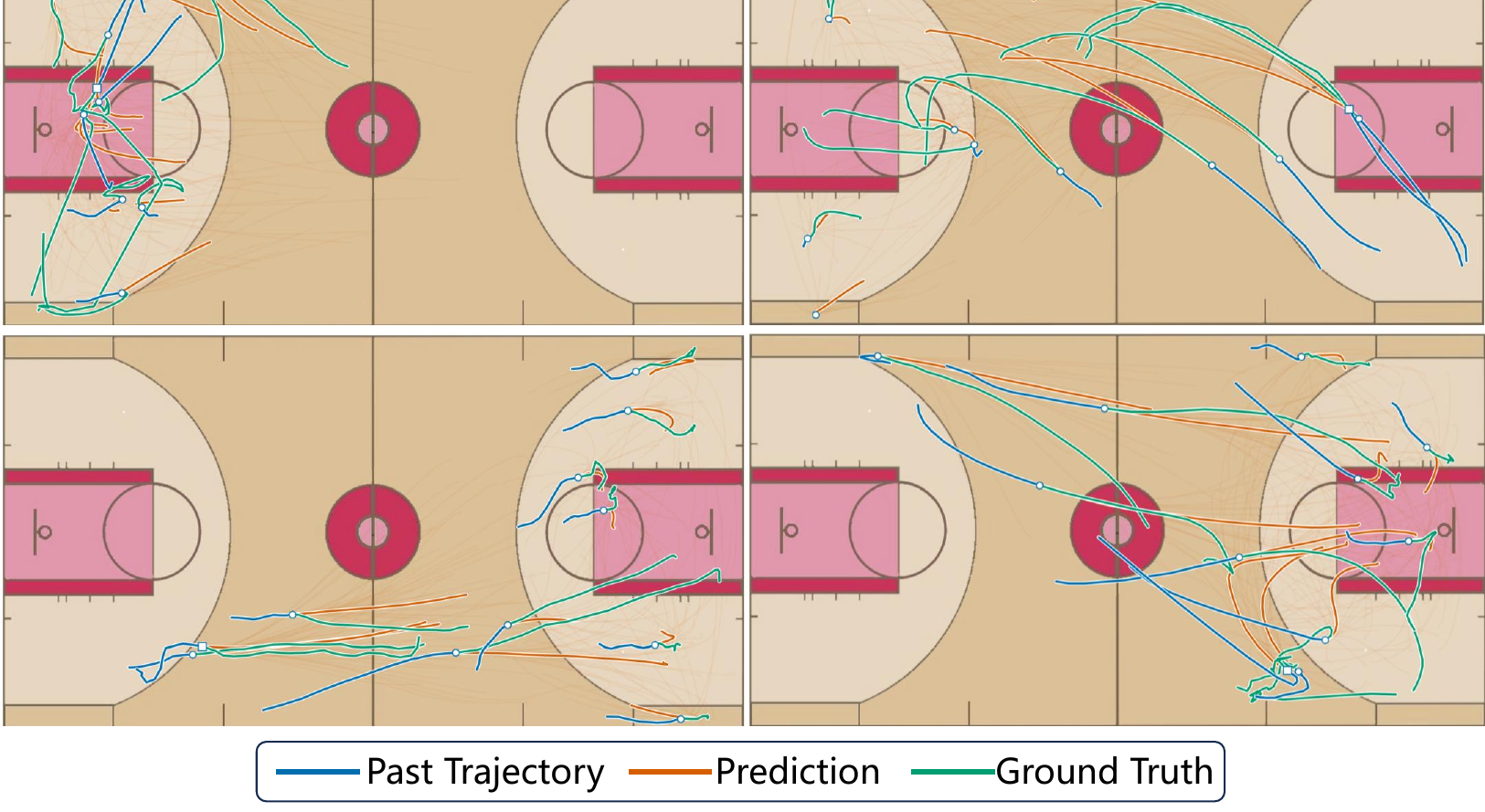}
\caption{Observed trajectories for ten players and the ball are input to \method. The model outputs a joint forecast. Ground-truth futures are shown for comparison.}
\label{fig:qualitative-nba}
\end{figure}

\begin{table}[t]
\centering
\small
\begin{tabular*}{\columnwidth}{@{\extracolsep{\fill}}lccc@{}}
\toprule
Method & Year & minADE & minFDE \\
\midrule
Social-GAN & 2018 & 27.23 & 41.44 \\
SoPhie & 2019 & 16.27 & 29.38 \\
PECNet & 2020 & 9.96 & 15.88 \\
PCCSNet & 2021 & 8.62 & 16.16 \\
Social-VAE & 2022 & 8.10 & 11.72 \\
MID & 2022 & 7.61 & 14.30 \\
LED & 2023 & 8.48 & 11.66 \\
TUTR & 2023 & 7.76 & 12.69 \\
ET+HighGraph & 2024 & 7.81 & \secondbest{11.09} \\
PPT & 2024 & \secondbest{7.03} & \textbf{10.65} \\
MoFlow & 2025 & 7.50 & 11.96 \\
IAD & 2026 & \textbf{6.85} & 11.22 \\
\midrule
\method & 2027 & 7.38 & 11.57 \\
\bottomrule
\end{tabular*}
\caption{Best-of-20 SDD results in pixels. Values follow IAD, with ET+HighGraph added from MoFlow~\cite{fu2025moflow,liu2026iad}.}
\label{tab:sdd}
\end{table}

\begin{table*}[t]
\centering
\small
\setlength{\tabcolsep}{5.5pt}
\begin{tabular*}{\textwidth}{@{\extracolsep{\fill}}lccc@{}}
\toprule
Variant & ETH/UCY & NBA & SDD \\
\midrule
\multicolumn{4}{@{}l}{\textit{Representation}}\\
Frame-wise position & 0.18/0.28 & 0.7310/0.9180 & 7.92/12.45 \\
Position DCT spectrum & 0.16/0.25 & 0.7160/0.8980 & 7.66/12.03 \\
Time-domain acceleration & 0.16/0.24 & 0.7110/0.8910 & 7.61/11.94 \\
\midrule
\multicolumn{4}{@{}l}{\textit{Phase-space supervision}}\\
Position only & 0.17/0.25 & 0.7160/0.8970 & 7.71/12.06 \\
Position + velocity & 0.16/0.23 & 0.7050/0.8810 & 7.52/11.81 \\
Position + acceleration & 0.16/0.24 & 0.7080/0.8860 & 7.58/11.90 \\
\midrule
\multicolumn{4}{@{}l}{\textit{Attention and endpoint conditioning}}\\
No hypothesis-axis attention & 0.17/0.26 & 0.7290/0.9150 & 7.81/12.31 \\
No agent-axis attention & 0.17/0.27 & 0.7180/0.8990 & 7.76/12.22 \\
No attention on either axis & 0.19/0.30 & 0.7480/0.9470 & 8.04/12.73 \\
No endpoint module & 0.17/0.26 & 0.7120/0.8950 & 7.86/12.39 \\
\midrule
\textbf{Full \method} & \textbf{0.15/0.22} & \textbf{0.6976/0.8694} & \textbf{7.38/11.57} \\
\bottomrule
\end{tabular*}
\caption{Component ablations reported as minADE/minFDE. All variants use best-of-20 sampling.}
\label{tab:ablation}
\end{table*}

\paragraph{SDD.}
\method obtains 7.38/11.57 on SDD. It ranks third in minADE and fourth in minFDE. The gaps to the best entries are 0.53 and 0.92 pixels. \method uses trajectories without scene images or semantic maps. Its result remains competitive, but visual context can still help resolve scene-dependent destinations.

The SDD ranking also clarifies the boundary of motion-only conditioning. Relative to the best entries, the ADE and FDE gaps are approximately 7.7\% and 8.6\%, respectively. In contrast, \method improves over MoFlow by 0.12 ADE and 0.39 FDE under the same trajectory-only input type. The remaining difference to PPT and IAD is therefore consistent with a dataset in which curbs, entrances, and walkable regions provide information that cannot always be inferred from eight observed positions. This comparison motivates adding scene semantics as a complementary condition rather than changing the acceleration-spectrum representation.

\subsection{Qualitative Results}

Figures~\ref{fig:qualitative-eth-sdd} and~\ref{fig:qualitative-nba} show forecasts on the three datasets. The ETH/UCY and SDD samples continue the observed heading and include curved paths. Each NBA sample contains all ten players and the ball, rather than independently selected agent paths. The examples show forecast geometry and grouping. They do not measure diversity, calibration, or likelihood.

\subsection{Ablation Studies}

Table~\ref{tab:ablation} isolates the main choices in the representation, objective, and architecture. The acceleration DCT spectrum gives lower error than either time-domain acceleration or the position DCT spectrum on all three benchmarks. Position, velocity, and acceleration supervision perform best when used together. Removing either attention axis increases error, and removing both produces the largest change within that block. A separately trained model without endpoint prediction and conditioning also performs worse than the full model.

\paragraph{Representation and supervision.}
Replacing frame-wise position generation with the full acceleration-spectrum model reduces ETH/UCY ADE/FDE from 0.18/0.28 to 0.15/0.22, reductions of 16.7\% and 21.4\%. The corresponding relative reductions are 4.6\%/5.3\% on NBA and 6.8\%/7.1\% on SDD. Time-domain acceleration is consistently better than frame-wise position, but remains worse than its DCT counterpart. The gain is therefore not explained by differentiation alone: concentrating motion in a truncated low-frequency basis also matters. Position DCT improves over frame-wise positions, yet acceleration DCT is better on every dataset, supporting the choice to place the generative state at the highest explicitly supervised kinematic order.

The phase-space rows show complementary supervision rather than three interchangeable losses. Adding velocity to position-only training reduces FDE from 0.25 to 0.23 on ETH/UCY, from 0.8970 to 0.8810 on NBA, and from 12.06 to 11.81 on SDD. Position plus acceleration is also better than position alone, but is slightly weaker than position plus velocity. The full objective improves again to 0.22, 0.8694, and 11.57 FDE. Velocity supplies a direct constraint on local displacement, while acceleration penalizes changes in that displacement; their combination is consistent with the deterministic two-stage reconstruction.

\paragraph{Interaction, modes, and endpoints.}
Removing both attention axes produces the largest architectural degradation: relative to the full model, FDE rises by 36.4\% on ETH/UCY, 8.9\% on NBA, and 10.0\% on SDD. Retaining only agent-axis attention is better on NBA and SDD than retaining only hypothesis-axis attention, which is expected for joint forecasts containing interacting entities. On ETH/UCY, the two single-axis variants are close, indicating that mode coordination and social interaction both contribute. The endpoint ablation raises FDE from 0.22 to 0.26, from 0.8694 to 0.8950, and from 11.57 to 12.39. Its larger relative effect on ETH/UCY and SDD is consistent with destination ambiguity in pedestrian scenes; on NBA, player interactions and short-horizon dynamics provide additional constraints.

\paragraph{Sampling efficiency.}
The supplement reports representation, bandwidth, classification, sample-count, endpoint-query, input-feature, and solver experiments, and separates a retrained endpoint ablation from an inference-time intervention. Reducing Euler integration from 100 to 20 steps preserves all three headline results at two-decimal precision. On NBA, latency falls from 17.55 to 3.68 ms per scene, or to approximately 21\% of the 100-step cost. Higher-order solvers yield only small raw-error changes while increasing runtime. The main tables retain the conservative 100-step setting for comparability, but the step ablation indicates that the learned vector field is smooth enough for substantially cheaper deployment.

\subsection{Design Trade-offs}

\paragraph{Spectral bandwidth.}
The bandwidth sweep retains one quarter, one half, three quarters, or all DCT coefficients. The half-spectrum setting used by the full model is best on every dataset: 0.15/0.22 on ETH/UCY, 0.6976/0.8694 on NBA, and 7.38/11.57 on SDD. A quarter spectrum raises FDE to 0.29, 0.9090, and 12.62, indicating that aggressive truncation removes trajectory changes needed for accurate endpoints. Retaining the full spectrum also degrades FDE to 0.24, 0.8820, and 11.78. The consistent U-shaped pattern supports a bias--variance interpretation: low frequencies capture the dominant motion, whereas some higher-frequency coefficients model turns and interaction responses; retaining every coefficient also exposes the flow to high-frequency variation that contributes less to displacement accuracy.

\paragraph{Hypothesis and endpoint budgets.}
Increasing the number of joint hypotheses from 1 to 5, 10, and 20 monotonically improves all three benchmarks. On NBA, ADE/FDE falls from 0.9210/1.3210 to 0.7620/0.9840, 0.7210/0.9050, and 0.6976/0.8694, while measured runtime grows from $1.0\times$ to $1.3\times$, $1.6\times$, and $2.0\times$. The largest gain occurs between one and five hypotheses, but the remaining increments are material under best-of-20 evaluation. Endpoint queries show clearer saturation. Increasing $L$ from 5 to 10 and 20 improves NBA FDE from 0.8970 to 0.8830 and 0.8694, whereas 40 queries gives 0.8700. ETH/UCY likewise stays at 0.15/0.22 for 20 and 40 queries, and SDD changes from 7.38/11.57 to 7.40/11.61. We therefore use 20 endpoint candidates: it captures the measurable gain without doubling a proposal set that already saturates.

\paragraph{Observed context.}
The observed-input sweep separates geometry from dynamics. Absolute positions alone obtain 0.19/0.31 on ETH/UCY, 0.7510/0.9540 on NBA, and 8.21/13.04 on SDD. Adding coordinates relative to the last observation improves all three; adding finite-difference velocity produces the full-model results. The same ordering across coordinate systems indicates that the gain is not tied to a benchmark-specific unit. Relative positions remove global offsets, while velocity exposes the local trend that must be continued or changed by the predicted acceleration. The hypothesis-classification head provides a smaller but consistent improvement: removing it changes the three FDE values from 0.22, 0.8694, and 11.57 to 0.24, 0.8860, and 11.88. This head does not create additional samples; it helps organize and select the existing joint hypotheses during training.

\paragraph{Endpoint diagnostic.}
Separately trained models without endpoint conditioning obtain 0.17/0.26, 0.7120/0.8950, and 7.86/12.39, while replacing predicted endpoints with ground-truth endpoints yields diagnostic upper bounds of 0.13/0.18, 0.6650/0.8110, and 6.92/10.54. The gap between predicted and oracle conditioning shows that endpoint estimation remains a meaningful error source, especially on SDD. A fixed-checkpoint NBA intervention is more conservative: bypassing the predicted endpoint changes 0.6976/0.8694 to 0.7052/0.8903. Together, the retraining and intervention tests separate two effects. Endpoint supervision shapes the learned representation over training, and the endpoint embedding also contributes directly at inference. The oracle row is not a test-time result and is used only to estimate remaining headroom.

\subsection{Cross-Dataset Analysis}

The three benchmarks use the same acceleration-spectrum target and reconstruction but differ in horizon, coordinate system, scene content, and number of agents. Their raw errors are therefore not directly comparable. On ETH/UCY, \method improves both metrics in every scene, with larger margins in FDE. On NBA, it is not best at every intermediate horizon, but gives the lowest four-second ADE and a tied-best FDE. Its smaller ranking advantage on SDD is consistent with the absence of visual scene cues, although the benchmark does not isolate that factor.

Best-of-20 metrics measure whether the sample set contains a trajectory near the ground truth. They do not measure calibration or likelihood. Results that become equal after rounding are treated as ties. The main experiments use one seed, so they do not quantify variation across training runs. The ablations compare controlled variants of \method, but four-decimal values should not be compared with baseline values reported at lower precision.

Taken together, the experiments support three distinct claims. The main tables establish competitive sample coverage across pedestrian, sports, and aerial-view benchmarks. The representation and objective ablations attribute the gain to acceleration-spectrum generation with joint kinematic supervision. The solver study shows that the result is not dependent on a finely discretized 100-step trajectory. They do not establish calibrated probabilities, robustness across training seeds, or the contribution of visual semantics. Those questions require likelihood- or calibration-aware metrics, repeated training, and controlled scene-feature experiments.

\section{Conclusion}

\method applies conditional flow matching to DCT coefficients of future acceleration. Fixed reconstruction links acceleration, velocity, and position to the observed initial state. Endpoint conditioning represents destination uncertainty. Alternating attention exchanges information among hypotheses and agents. The model gives the best reported results on ETH/UCY and four-second NBA ADE, ties the best NBA FDE, and remains competitive on SDD.

Despite its strong performance across diverse motion regimes, SpecFlow remains a motion-only forecaster: it does not incorporate scene imagery, walkable-region priors, or semantic destination cues. The comparatively smaller gains on SDD therefore highlight the complementary role of scene context in resolving destination ambiguity, rather than a limitation of the acceleration-spectrum representation itself. Moreover, best-of-20 displacement metrics evaluate sample-set coverage but do not characterize probability calibration, likelihood quality, or the distribution of generated modes. Extending the proposed structured generative state with scene-aware conditioning and calibration-sensitive evaluation constitutes a natural direction for future work.

\clearpage
\bibliography{aaai2027}

\end{document}